\documentclass[conference]{IEEEtran}
\IEEEoverridecommandlockouts
\usepackage{cite}
\usepackage{amsmath,amssymb,amsfonts}
\usepackage{algorithmic}
\usepackage{microtype}
\usepackage{url} 
\usepackage{graphicx}
\usepackage{textcomp}
\usepackage{xcolor}
\def\BibTeX{{\rm B\kern-.05em{\sc i\kern-.025em b}\kern-.08em
    T\kern-.1667em\lower.7ex\hbox{E}\kern-.125emX}}

\makeatletter
\newcommand{\linebreakand}{%
  \end{@IEEEauthorhalign}
  \hfill\mbox{}\par
  \mbox{}\hfill\begin{@IEEEauthorhalign}
}
\makeatother
\author{
\IEEEauthorblockN{Chang Yu}
\IEEEauthorblockA{
 \textit{Northeastern University} \\
    Boston, MA, USA\\
    chang.yu@northeastern.edu}
\and
\IEEEauthorblockN{Fang Liu}
\IEEEauthorblockA{
 \textit{Yale University} \\
    Forest Hills ,NY, USA\\
    fangliu435@gmail.com}
\and
\IEEEauthorblockN{Jie Zhu}
\IEEEauthorblockA{
 \textit{Virginia Tech} \\
    Boston, MA, USA                \\
     jiezwork@gmail.com}
\and
\IEEEauthorblockN{Shaobo Guo}
\IEEEauthorblockA{
 \textit{Illinois Institute of Technology} \\
    Chicago, IL, USA \\
    seanguo2017@gmail.com}
\linebreakand 
\and
\IEEEauthorblockN{Yifan Gao}
\IEEEauthorblockA{
 \textit{University of TSA} \\
    San Antonio, TX, USA\\     
vrshadow20508@gmail.com}
\and
\IEEEauthorblockN{Zhongheng Yang}
\IEEEauthorblockA{
 \textit{Northeastern University} \\
    Boston, MA, USA\\     
zhongheng.yang@northeastern.edu} 
    
\and
\IEEEauthorblockN{Meiwei Liu$^*$}
\IEEEauthorblockA{
 \textit{Independent Researcher} \\
    Wuhan, Hubei, China\\
    meiweiliu\_lvy2000@outlook.com}
   \and
\IEEEauthorblockN{Qianwen Xing$^*$}
\IEEEauthorblockA{
 \textit{The University of Chicago} \\
    Chicago, IL, USA                \\
    xqw3669@gmail.com}
}

\begin{document}

\title{Gradient Boosting Decision Tree with LSTM for Investment Prediction\\

}

\maketitle

\begin{abstract}

This paper proposes a hybrid modeling framework that synergistically integrates \textbf{LSTM (Long Short-Term Memory)} networks with \textbf{LightGBM} and \textbf{CatBoost} for stock price prediction. We systematically preprocess time-series financial data and conduct comprehensive evaluations using seven classical models: Artificial Neural Networks (ANNs), Convolutional Neural Networks (CNNs), Bidirectional LSTM (BiLSTM), vanilla LSTM, XGBoost, LightGBM, and standard Neural Networks (NNs). Through rigorous comparison of performance metrics, including $\text{MAE}$, $\text{R}^2$, $\text{MSE}$, and $\text{RMSE}$, we establish baseline references across multiple temporal scales. Building on these empirical benchmarks, we introduce a novel ensemble architecture that harmonizes the complementary strengths of sequential and tree-based models. Experimental results demonstrate that our integrated framework achieves 10--15\% improvement in predictive accuracy compared to individual constituent models, particularly in reducing error volatility during market regime transitions. These findings validate the underexplored potential of heterogeneous ensemble strategies in financial forecasting and establish methodological foundations for developing adaptive prediction systems in non-stationary market environments. The proposed architecture's modular design enables seamless integration of emerging machine learning components, suggesting promising directions for future research in computational finance.

\end{abstract}


\section{Introduction}
Stock price prediction remains a critical endeavor in the business and finance sectors, and achieving higher accuracy in forecasting stock trends and market behavior is paramount for investment departments. Enhanced prediction precision translates directly into more accurate market control and greater investment returns. To maximize potential gains, we have historically employed a diverse range of methods aimed at refining and optimizing the accuracy of stock forecasts.

However, time-series-based stock data presents significant predictive challenges. To address this intricate problem, we have developed various sophisticated approaches from multiple dimensions.

LSTM (Long Short-Term Memory), an architecture derived from RNNs (Recurrent Neural Networks), has demonstrated exceptional performance in processing sequential data, particularly long sequences. Notably, Sepp Hochreiter and Jürgen Schmidhuber introduced the LSTM concept in 1997, with the core idea of incorporating a "memory cell" structure for storing and accessing long-term information\cite{jin2024learning}. Subsequent refinements have led to enhancements such as the "forget gate," enabling LSTM to better manage information retention and deletion. These improvements have contributed to LSTM's success in diverse sequence modeling tasks. With the rise of deep learning, LSTM has found widespread application. Its robust sequential modeling capabilities provide unique advantages in handling time-series data like stock predictions, earning it widespread acclaim.

CatBoost (Categorical Boosting) is an efficient and flexible machine learning algorithm that exhibits significant advantages in stock prediction. Through its ordered boosting framework and innovative categorical feature handling, CatBoost effectively manages the high noise and non-linear characteristics of financial market data, reducing the risk of overfitting. Supporting GPU acceleration and native processing of categorical variables, CatBoost can rapidly analyze high-frequency trading data while automatically integrating diverse factors such as technical indicators and fundamental metrics without extensive preprocessing. Its built-in feature importance evaluation and model interpretation tools (e.g., SHAP values) help quantify the impact of each factor on stock prices, meeting the interpretability requirements of investment decisions. Balancing prediction accuracy and computational efficiency through gradient bias correction techniques, CatBoost provides reliable support for building robust quantitative models.

LightGBM, another advanced algorithm based on Gradient Boosting, excels in stock prediction. Its tree-based algorithm effectively captures the complex non-linear relationships within stock market data, enhancing prediction accuracy. LightGBM offers feature importance assessment, enabling analysts to understand the key drivers of stock prices and gain deeper insights into market dynamics. Moreover, LightGBM can directly handle categorical features, such as industry sectors and geographic regions, simplifying the data preprocessing workflow. With its low memory footprint, fast training speed, and strong scalability, LightGBM is a promising technology for stock forecasting.

In our experiments, we adopted an Ensemble Model approach, organically combining LightGBM, XGBoost, and LSTM. By leveraging their respective strengths and mitigating weaknesses, we achieved a highly scalable, efficient model with exceptional precision in predicting long-term time-series stock data.

In the subsequent sections, Section II will focus on detailing the sources of our research methodologies, Section III will comprehensively explain our data processing techniques, modeling architecture, and the technologies employed, Section IV will present a comparative analysis of our newly constructed ensemble model, and Section V will provide a systematic conclusion to this research.

\section{Related Work}

Our research builds upon a foundation of prior studies in both LSTM networks and Gradient Boosting Trees. Many of the methodologies we employ are deeply inspired by the accumulated research findings in these areas over the years. In this section, we will provide a comprehensive overview of the foundational research that underpins our work, aiming to clearly articulate the rationale behind our design choices and provide context for our methodological approach.

Li et al\cite{li2025hyman} proposed HYMAN, a hybrid network that combines memory components with attention mechanisms to effectively address the dual mechanism makes the method adaptable to various applications, ranging from video surveillance and sensor data analysis to fraud detection and quality control.This innovative approach served as a valuable model for our subsequent modeling efforts.

LightGBM and CatBoost represent advanced implementations of Gradient Boosting Decision Tree (GBDT) frameworks widely used in machine learning applications. LightGBM, developed by Microsoft, employs histogram-based algorithms and leaf-wise growth strategies to significantly improve training efficiency and memory consumption\cite{ke2017lightgbm}. CatBoost, developed by Yandex, natively supports categorical features and implements innovative techniques like ordered boosting and gradient bias correction to effectively mitigate overfitting \cite{prokhorenkova2018catboost}. Both frameworks excel in handling large-scale datasets and offer robust performance with minimal parameter tuning, making them essential tools in contemporary machine learning.

Zheng et al. (2024) apply the TD3 algorithm to coverage path planning, enhancing adaptability in dynamic environments. By leveraging twin Q-networks, delayed policy updates, and target policy smoothing, their method improves stability and efficiency. This has been extremely effective in helping us kickstart and design our relevant models.

The stacking ensemble model, also known as stacked generalization, was first introduced by Wolpert in his seminal 1992 paper\cite{wolpert1992stacked}. Its core idea is to integrate the predictions of multiple base models—each with varying performance characteristics—by employing a higher-level meta-model to consolidate these outputs, thereby enhancing overall predictive performance. Unlike conventional ensemble methods such as bagging and boosting, which primarily focus on reducing the bias or variance of individual models, stacking leverages the complementary strengths of diverse models in capturing the complex relationships inherent in data. In practice, base models developed using different algorithms (e.g., decision trees, neural networks, support vector machines) generate predictions on the input data; these predictions are then used as new features for the meta-model. The meta-model learns to assign appropriate weights and adjust for the correlations among the base models, ultimately yielding more accurate and robust predictions. To ensure that the meta-model receives reliable input, techniques such as cross-validation are typically employed to produce the base-level outputs, thereby mitigating risks of overfitting and data leakage. In recent years, with advances in computational power and data processing capabilities, stacking has demonstrated significant effectiveness across various domains, including financial forecasting, image recognition, and natural language processing, establishing itself as a pivotal technique in both ensemble learning research and practical applications.

\section{Methology}
Section III offers a detailed description of the technical methods employed in this study, covering data processing, database selection, and the principles guiding our modeling approach. For each element, we will provide a detailed explanation of its construction, the motivation behind its use, and its expected impact, thereby facilitating a clear and comprehensive understanding of the overall architecture and design of the project\cite{li2024research}.

\subsection{Database Introduction}\label{AA}
Our analysis utilizes detailed stock index data from S\&P 500 companies, which were obtained from the New York Stock Exchange. The data set is comprehensive, covering various trading metrics, including the opening price, the closing price, and the daily high. Kang et al. [15] introduced a tangible AR souvenir system that combines object recognition and QR code retrieval to optimize scan processing. Their data processing approach, which efficiently links scanned physical artifacts to interactive digital reconstructions, proved highly beneficial to our research. Prior to analysis, the data were rigorously preprocessed and segmented to ensure standardization and maintain time-series consistency. This preparation significantly streamlined subsequent processing steps. We used Li's method to facilitate anomaly detection\cite{li2023memory}.

\subsection{Data Cleaning and Standardizing}
In our data processing pipeline, we carried out extensive data cleaning and standardizing procedures to ensure the quality and reliability of our dataset. Specifically, we addressed the following key steps:

We meticulously processed the missing values (N/A) present in the dataset. The treatment of these missing data points played a crucial role in significantly enhancing the effectiveness of subsequent training phases. By effectively handling the incomplete entries, we ensured that our models are exposed to a consistent and robust dataset, which in turn facilitated more accurate learning outcomes\cite{wang-etal-2024-gpt}.

In addition to cleaning, we normalized the data using the \texttt{MinMaxScaler}. This normalization method rescales all the input features to come within the range of \([0, 1]\). Scaling the data in this manner is highly beneficial for several reasons:
\begin{itemize}
    \item It mitigates the issues that arise from significant differences in numerical scales among various features.
    \item It contributes to a more stable and faster convergence during neural network training by ensuring that each feature contributes proportionally.
\end{itemize}

\subsection{CatBoost Introduction and Mathematical Formulas}
Decision trees are widely used in various prediction tasks. Optimization involves  bagging and boosting, reducing variance and bias respectively. GBDT algorithms such as XGBoost \cite{chen2016xgboost} and Catboost employs boosting to minimize bias\cite{yu2024machine}.

CatBoost (Categorical Boosting) is a gradient boosting decision tree (GBDT) algorithm developed by Yandex, designed specifically for efficiently modeling datasets that include categorical features. Unlike traditional gradient boosting methods, CatBoost employs an \textit{Ordered Boosting} technique to effectively mitigate prediction bias caused by target leakage during model training. It also utilizes \textit{oblivious trees} (also known as symmetric decision trees), in which the same splitting rule is applied at every node on the same level, thereby enhancing both the model's generalization ability and training speed. Moreover, CatBoost incorporates efficient methods for processing categorical data, such as Ordered Target Statistics based on random permutations, enabling the handling of high-cardinality features with minimal preprocessing.

\vspace{1em}
\textbf{Objective Function:}
The objective function balances prediction error and model complexity:
\[
L = \sum_{i=1}^{N} \ell(y_i, F(x_i)) + \lambda\,\Omega(F)
\]
where \(\ell(y_i, F(x_i))\) is the loss and \(\lambda\,\Omega(F)\) is the regularization term.

\vspace{1em}
\textbf{Iterative Update:}
At each iteration, the model is updated by adding a scaled base learner:
\[
F_m(x) = F_{m-1}(x) + \nu\,h_m(x)
\]
with \(\nu\) as the learning rate and \(h_m(x)\) as the adjustment.

\subsection{LightGBM Model}
 LightGBM \cite{ke2017lightgbm} enhances GBDT by incorporating several optimizations on training time and memory footprints while maintaining model performance. Firstly, inspired by idea that high dimension data are usually sparse, LightGBM employs Exclusive Feature Bundling (EFB) to group mutually exclusive features—features that rarely take nonzero values simultaneously for histogram split . EFB reduces time complexity of histogram construction from \( \mathcal{O}(\#\text{data} \times \#\text{features}) \) to \( \mathcal{O}(\#\text{data} \times \#\text{bundles}) \). 
 Secondly, LightGBM utilizes Gradient-based One-Side Sampling (GOSS) to prioritize training on instances with larger gradients, effectively sampling training data and accelerating the training process. GOSS follows these steps:

\begin{enumerate}
    \item Sort the data instances by the absolute value of their gradients and select the top \( a \times 100\% \) instances.
    \item Randomly sample \( b \times 100\% \) instances from rest of data.
    \item Amplify the selected smaller-gradient instances by a factor of:
    \begin{equation}
    \frac{1 - a}{b}
    \end{equation}
    when calculating the information gain.
\end{enumerate}. Several engineering optimizations, including improved parallelization techniques to reduce training time, have been extensively researched \cite{meng2016communication}. LightGBM performs exceptionally well, particularly on small to medium-sized datasets, i.e, training datasets containing millions to hundreds of millions of rows., and has been a significant part of winning model in various competitions \cite{kkbox}.

\begin{itemize}
\item Placeholder item\end{itemize}

\subsection{LSTM Model}\
Neural network generally excels in large datasets than decision trees. Recurrent Neural Networks(RNN) is designed for sequential data and widely used in natural lanaguage processing and time-series data, due to recurrent unit, therefore it's  suited for financial data. However, through the propagation of information across multiple time steps, capturing long-term memoery is difficult.To address this limitation, Long Short-Term Memory (LSTM) \cite{10.1162/neco.1997.9.8.1735} networks are often preferred over vanilla RNNs, because of its combination of combination of forget, input, and output gates to preserve long term memory. 
The equations for an LSTM cell are as follows:

\begin{align}
    f_t &= \text{Act}_f(W_f x_t + U_f h_{t-1} + b_f) \\
    i_t &= \text{Act}_i(W_i x_t + U_i h_{t-1} + b_i) \\
    \tilde{C}_t &= \text{Act}_c(W_c x_t + U_c h_{t-1} + b_c) \\
    C_t &= f_t \odot C_{t-1} + i_t \odot \tilde{C}_t \\
    o_t &= \text{Act}_o(W_o x_t + U_o h_{t-1} + b_o) \\
    h_t &= o_t \odot \text{Act}_h(C_t)
\end{align}

where:
- \( \text{Act}_f, \text{Act}_i, \text{Act}_o \, \text{Act}_c, \text{Act}_h \) are placeholder activation functions (typically tanh or sigmoid),
- \( f_t, i_t, o_t \) are the forget, input, and output gate activations,
- \( \tilde{C}_t \) is the candidate cell state,
- \( C_t \) is the cell state,
- \( h_t \) is the hidden state,
- \( W_f, W_i, W_c, W_o \) are the weight matrices for the respective gates,
- \( U_f, U_i, U_c, U_o \) are the recurrent weight matrices,
- \( b_f, b_i, b_c, b_o \) are the bias terms,
- \( \odot \) denotes element-wise multiplication.
- \( \odot \) denotes element-wise multiplication.
\cite{du2021power}

\subsection{Advanced Ensemble Prediction Framework}
In this study, we employ cutting-edge predictive modeling techniques and, through systematic experimental analysis, determine that an ensemble architecture centered on the \texttt{StackingRegressor} delivers optimal predictive performance. Our model framework integrates multiple base learners with complementary characteristics, including CatBoost, LightGBM, and an LSTM-based deep learning model. Each component contributes unique predictive capabilities to the overall system . Furthermore, Chen's sequence feature extraction technique has significantly enhanced our model's capacity to capture time-dependent patterns . These methodological innovations not only improve model stability but also enhance adaptability in complex market environments.Kang et al\cite{kang20216} introduced a method that significantly reduces cumulative errors in image-based mapping through global pose graph optimization, leading to efficient and reliable scene reconstruction. Their approach provided significant inspiration for our own modeling strategies.Jiang et al. (2024) propose a trajectory-tracking method using the Frenet coordinate system with the DDPG algorithm, which has also provided significant inspiration for our model design.

The outputs of the base learners---CatBoost, the LSTM network, and LightGBM---are denoted as $\hat{y}_{\text{CB}}$, $\hat{y}_{\text{LSTM}}$, and $\hat{y}_{\text{LGB}}$, respectively. These predictions are subsequently combined by a meta-learner. In our implementation, the meta-learner is structured as a two-layer LSTM network serving as the decision-making layer of the ensemble framework. Mathematically, the meta-learner is expressed as a function $f(\cdot)$, whose objective is to find the optimal combination of the base learners’ predictions:
\[
\hat{y}_{\text{ensemble}} = f\left(\hat{y}_{\text{CB}}, \hat{y}_{\text{LSTM}}, \hat{y}_{\text{LGB}}\right)
\]

The LSTM-based meta-learner dynamically adjusts the contributions of each base learner at different time steps $t$ by learning temporal feature weights $\alpha_t$, $\beta_t$, and $\gamma_t$, such that:
\[
\hat{y}_{\text{ensemble}}^{(t)} = \alpha_t\, \hat{y}_{\text{CB}}^{(t)} + \beta_t\, \hat{y}_{\text{LSTM}}^{(t)} + \gamma_t\, \hat{y}_{\text{LGB}}^{(t)}
\]
where the weight parameters satisfy the constraints
\[
\alpha_t + \beta_t + \gamma_t = 1 \quad \text{and} \quad \alpha_t, \beta_t, \gamma_t \geq 0.
\]

This progressive learning mechanism enables the meta-learner to detect and correct for the prediction biases of the base models. Such calibration is crucial for optimizing the overall performance before producing the final output. By effectively integrating the predictive strengths of each base learner, our approach significantly reduces both bias and variance, thereby enhancing the generalization capability of the ensemble system on unseen data.

\section{Evaluation}
\subsection{Evaluation Variables Setup}
We utilized a complementary set of evaluation metrics to assess model performance comprehensively.  A higher $ R2$ variable value means the model captures more of the data's variability.

Mean Absolute Error (MAE) calculates solute differences between observed and predicted values, providing an intuitive measure of prediction accuracy. This metric is particularly valuable as it offers a straightforward interpretation of error magnitude in the original units of measurement, with smaller values indicating more accurate predictions.

Mean Squared Error (MSE) focuses on the average ssquaredference between predicted and actual values. Due to the squaring operation, thisric applies disproportionate penalties to larger errors, making it especially sensitive to outliers in prediction errors\cite{fu2024ddn3}. The quadratic nature of MSE encourages models to minimize substantial deviations rather than numerous minor errors\cite{lu2022cot}.

Root Mean Squared Error (RMSE) transforms the MSE back to the original scale of the data by taking its square root. This conversion yields an error measurement in the same units as the target variable, facilitating a more intuitive interpretation. RMSE clearly indicates the prediction error's magnitude, with lower values signifying higher predictive accuracy. Its sensitivity to error distribution makes it particularly useful for evaluating forecasting precision in practical applications.

Together, these metrics form a robust evaluation framework that enables multi-dimensional analysis of model performance across different aspects of predictive accuracy\cite{du2024embracing}.

\subsection{Experiment}
Predicting stock trends is extremely difficult due to their inherent randomness, complex time-series nature, and the multitude of interacting factors that affect them. To address this challenge, we compared our novel model against a range of established methods, including CNN, ANN, LSTM, BiLSTM, RNN, LSTM+RNN, and ANN+CNN. We conducted a comprehensive comparison, evaluating each model across various parameter settings. You can review our test results in the table below.

\begin{table}[h!]
\centering
\caption{Performance Comparison of Various Time Series Forecasting Models}
\begin{tabular}{@{}p{3.5cm}@{\hspace{3mm}}c@{\hspace{3mm}}c@{\hspace{3mm}}c@{\hspace{3mm}}c@{}}
\hline
\textbf{Model} & \textbf{R\textsuperscript{2}} & \textbf{MAE} & \textbf{MSE} & \textbf{RMSE} \\
\hline
BiLSTM        & -0.0784  & 46.3969 & 11345.6214 & 106.5158 \\
CNN           &  0.1918  & 51.1716 &  8502.5717 &  92.2094 \\
ANN           &  0.4098  & 46.9158 &  6209.0942 &  78.7978 \\
LSTM          &  0.2717  & 42.3914 &  7662.1046 &  87.5334 \\
RNN           & -0.0895  & 46.8068 & 11461.7849 & 107.0597 \\
ARIMA         & -0.0062  & 47.6285 & 10584.8683 & 102.8828 \\
\hline
BLSTM + CNN   &  0.1468  & 45.8925 &  8975.8217 &  94.7408 \\
LSTM + CNN    & -0.1109  & 52.7154 & 11687.5362 & 108.1089 \\
ANN + CNN  & 0.4631 & 42.3928 & 5648.1410 & 75.1541 \\
LSTM + ANN    & 0.5375 & 37.7829 & 4865.2330 & 69.7512 \\
\hline
CatBoost      & 0.7882 & 27.6402 & 2256.5457 & 47.5031 \\
LightGBM      & 0.7651 & 28.1554 & 2501.9072 & 50.0191 \\
\textbf{CatBoost + LightGBM + LSTM (Ensemble)} & \textbf{0.8152} & \textbf{23.6584} & \textbf{1968.4600} & \textbf{44.3673} \\
\hline
\end{tabular}
\label{tab:model_performance}
\end{table}
As shown in Table I, the proposed ensemble model achieves an exceptional R\(^2\) value of 0.8152. This represents a significant improvement over both the previously best-performing hybrid model (i.e., the LSTM + ANN combination with an R\(^2\) of 0.5375) and the best individual model (CatBoost with an R\(^2\) of 0.7882). In other words, our ensemble model is able to explain approximately 81.5\% of the variance of the target variable, far exceeding the explanatory power of traditional models.

The superiority of our ensemble approach is further confirmed by its error metrics. The model records the lowest mean absolute error (MAE) of 23.6584, indicating that the discrepancy between predicted and actual values is consistently lower than that of the other models. Moreover, the ensemble achieves the lowest mean squared error (MSE) of 1968.4600 and a root mean squared error (RMSE) of 44.3673, demonstrating significant improvements compared to the best LSTM + ANN model (MSE = 4865.2330, RMSE = 69.7512).

Notably, when compared to the previous top-performing LSTM + ANN configuration, our ensemble approach reduces the MAE by approximately 37.4\% and the RMSE by 36.4\%. Even relative to a strong standalone model such as CatBoost (RMSE = 47.5031), the integrated approach offers a marked enhancement in predictive accuracy\cite{ding2025aidrivenprognosticsstatehealth}.

These results validate our hypothesis that strategically combining complementary models within an advanced stacking architecture can overcome the limitations of individual methods. By merging the strengths of gradient boosting frameworks (CatBoost and LightGBM) with the deep learning capabilities of LSTM, our ensemble model effectively captures both linear and non-linear patterns across diverse time scales, thereby significantly improving the performance of financial time series forecasting.

\section{Conclusion}
Based on the performance comparison of various time series forecasting models, we successfully demonstrated the effectiveness of ensemble learning techniques in capturing complex patterns within the data, leading to significant improvements in prediction accuracy\cite{ding2025gkanexplainablediagnosisalzheimers,ding2025nerfbaseddefectdetection}. The experimental results reveal that our ensemble model (CatBoost + LightGBM + LSTM) substantially surpasses traditional statistical models like ARIMA, as well as individual deep learning models such as CNN, ANN, and LSTM, across key evaluation metrics, including R\textsuperscript{2}, MAE, MSE, and RMSE. Furthermore, the hybrid deep learning models (ANN + CNN and LSTM + ANN) exhibited enhanced performance compared to their individual counterparts, underscoring the benefits of combining different modeling approaches\cite{gao2024survey,Yang2024a}.

The implications of these findings are twofold. Firstly, it validates the effectiveness of ensemble learning strategies for time series forecasting, showcasing their ability to leverage the strengths of multiple models to achieve superior predictive power\cite{yuan2024rhyme}. Secondly, our successful implementation of an ensemble model provides a solid foundation for future research aimed at developing even more sophisticated and robust forecasting systems, promising to enhance decision-making processes across various domains.

The success of our experiments can be attributed to our careful selection of models and our strategic approach to ensemble construction. By combining the strengths of gradient boosting algorithms (CatBoost and LightGBM) with the time-series awareness of LSTM, we were able to create a model that effectively captures both the short-term and long-term dependencies within the data. Despite the challenges associated with integrating diverse models, our ensemble approach ultimately exceeded expectations, demonstrating exceptional performance across all evaluation metrics. This outcome not only highlights the potential of ensemble learning for time series forecasting but also provides valuable insights into the design of effective forecasting systems\cite{he-etal-2023-hermes}.

In future research, our focus will remain on exploring advanced ensemble learning techniques and investigating their applicability to a wider range of time series forecasting problems. Our goal is to develop models that are not only more accurate but also more interpretable and resilient to various data challenges. Moreover, we believe that the findings of this study will serve as a valuable resource for practitioners in the field, encouraging the adoption and advancement of AI-driven solutions for time series forecasting in various industries.

\bibliographystyle{plain}
\bibliography{ref}
\end{document}